\documentclass{article} 
\usepackage{nips14submit_e,times}
\usepackage{hyperref}
\usepackage{url}
\usepackage{amstext}
\usepackage{amsmath}
\usepackage{amsfonts,amssymb}
\usepackage{graphicx}
\usepackage{array}
\usepackage{upgreek}

\usepackage{wrapfig}

\newcommand{\argmin}{\mathop{\mathrm{argmin}}}
\newcommand{\diag}{\mathop{\mathrm{diag}}}
\newcommand{\GR}{GR}
\newcommand{\LR}{LR}
\newcommand{\DR}{DR}
\newcommand{\Trans}{\intercal}

\title{Deep Regression for Face Alignment}

\author{
Baoguang Shi$^1$\thanks{This work was done while the author visiting Microsoft Research, Beijing, China.} ~~~~
Xiang Bai$^1$~~~~
Wenyu Liu$^1$~~~~
Jingdong Wang$^2$~~~~\\
{\small $^1$Dept. of Electronics and Information Engineering, Huazhong Univ. of Science and Technology, China} \\
{\small $^2$Microsoft Research, Beijing, China} \\
{\small\tt shibaoguang@gmail.com,\{xbai,liuwy\}@hust.edu.cn,jingdw@microsoft.com} \\
}

\nipsfinalcopy 

\begin{document}

\maketitle


\begin{abstract}
    In this paper, we present a deep regression approach for face alignment.
    The deep architecture consists of a global layer and multi-stage local layers.
    We apply the back-propagation algorithm with the dropout strategy to jointly optimize the regression parameters.
    We show that
    the resulting deep regressor gradually and evenly approaches the true facial landmarks
    stage by stage,
    avoiding the tendency to yield over-strong early stage regressors
    while over-weak later stage regressors.
    Experimental results show that our approach achieves the state-of-the-art performance on the benchmark datasets.
\end{abstract}


\section{Introduction}

Face alignment, a.k.a. facial landmark localization,
is a fundamental problem in computer vision.
It aims to predict landmark positions given a 2D facial image.
This problem has attracted a lot of research efforts~\cite{cootes1995active,cootes2001active,yuille1992feature,liang2008face,ding2008precise,belhumeur2011localizing,matthews2004active,cristinacce2007boosted}.
However, it remains challenging
when face images are taken under uncontrolled conditions with large variation on poses, expressions and lighting conditions.



Cascaded regression has achieved the state-of-the-art performance.
Cascaded pose regression \cite{dollar2010cascaded}
and the following work explicit shape regression \cite{cao2014face}
sequentially learn a cascade of random fern regressors using shape indexed features
and
progressively regress the shape stage by stage
over the learnt cascade.
Robust cascaded pose regression~\cite{burgos2013robust}
extends cascaded pose regression
with occlusion handling,
enhanced shape-indexed features and more robust initialization.
Supervised descent method \cite{xiong2013supervised}
shows that a cascade of simple linear regressors is able to achieve the superior performance.
Local binary feature regression~\cite{renface}
speeds up the supervised descent method
using the learned trees-induced binary feature representation.

We observe that
the cascaded regression approach tends to
learn over-strong early stage regressors
but over-weak later stage regressors.
The reason is that the multi-stage regressors
are learnt sequentially
from the first stage regressor
to the last stage regressor.
Inspired by the natural fact that
cascaded regression is a deep neural network,
we propose a deep regression approach
that adopts the back-propagation algorithm
with the dropout strategy
to jointly optimize a deep structure.
The resulting deep regressor gradually
and simultaneously reduces the bias and the variance
of the estimation from the first regressor
to the last regressor,
thus yielding a better facial landmark location.
The structure illustrated in Figure~\ref{fig:archoverview}.a
consists
of two sub-networks:
a global layer
and multi-stage local layers.
The latter sub-network is the same
to the structure of supervised decent method~\cite{xiong2013supervised},
and each local layer contains
a local feature extraction sub-layer
and a local regressor.
The former sub-network
aims to provide an initial result
regressed from the facial image
as the input of the latter local regressors.

There are some other attempts to adopt deep learning for face alignment.
A cascade of three convolutional neural network (CNN) regressors~\cite{sun2013deep}
each of which regresses the facial landmark positions
is used for face alignment.
Another deep learning solution,
coarse-to-fine CNN cascade~\cite{zhou2013extensive}
is developed for face alignment.
The two algorithms are different from our approach
as all the CNNs are trained separately,
in contrast our approach learns all-stage regressors jointly.
In essence,
the two algorithms can benefit from jointly optimizing all the three CNN regressors.


\section{The Architecture} \label{sec:arch}

Let the vector $\mathbf{s} = [ x_{1}, y_{1}, \dots, x_{P}, y_{P} ]^{\Trans} \in \Re^{2P}$
be the shape of the face,
where $(x_{p}, y_{p})$
is the position of the $p$-th landmark.
The task of face alignment is
to predict all the $P$ landmark positions,
i.e.,
the shape $\mathbf{s}$ from the facial image $I$.

\begin{figure}[t]
    \begin{centering}
    \includegraphics[width=\linewidth]{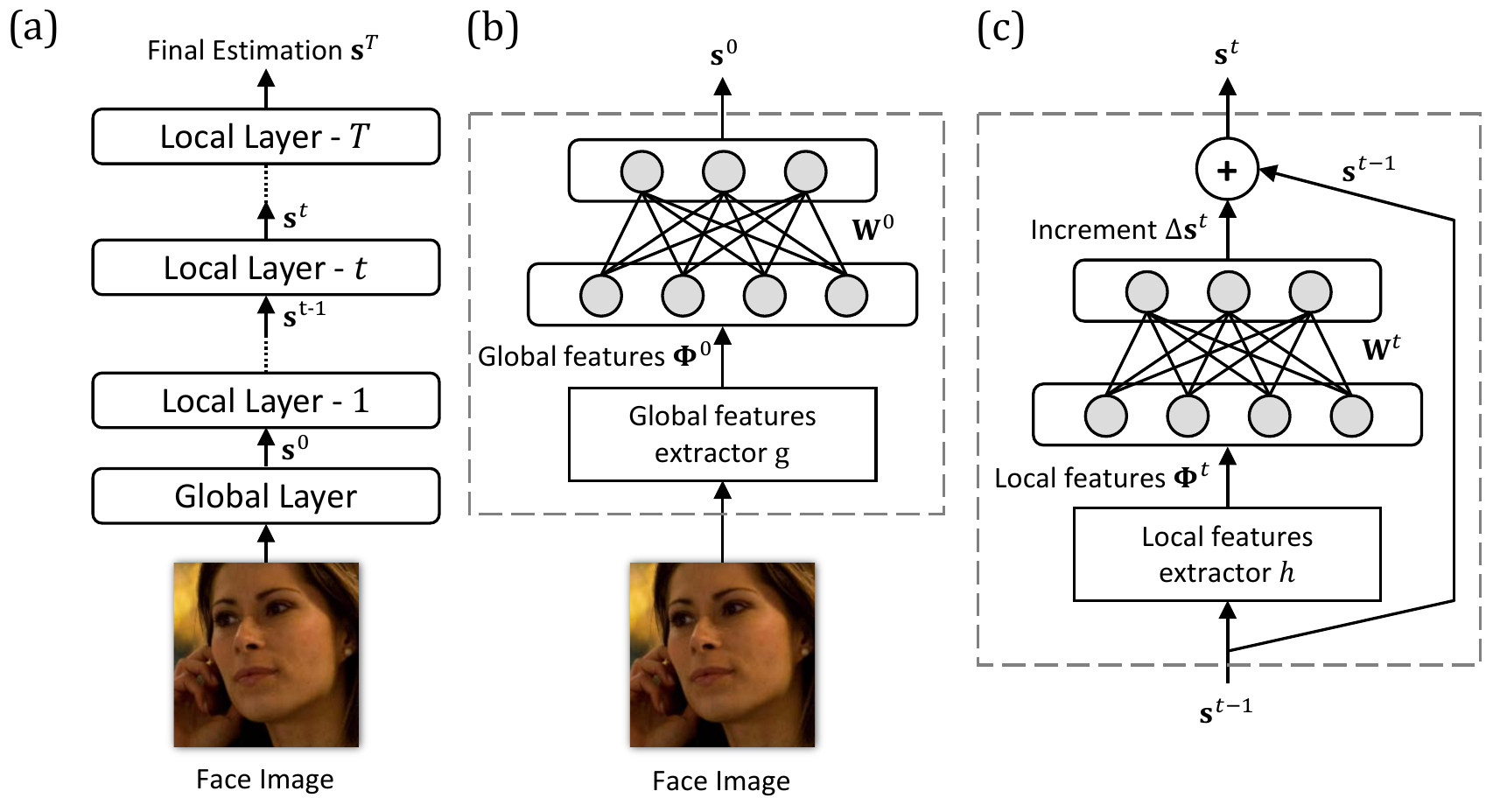}
    \par\end{centering}
    \caption{(a) Overview of the proposed learning architecture. The network takes face image as input and outputs shape estimation $\mathbf{s}^{T}$. The global layer estimates initial shape $\mathbf{s}^{0}$ and the rest local layers refine the estimation iteratively. (b) Inner structure of the global layer, see Section \ref{sec:globalreg} for details. (c) Inner structure of the $t$-th local layer, see Section \ref{sec:localreg} for details.}
    \label{fig:archoverview}
\end{figure}

The architecture
is a multi-layered deep network,
depicted in Figure \ref{fig:archoverview}.a.
The network consists of $1+T$ layers,
with the first global layer
and the rest $T$ local layers.
The global layer
consists of a global feature extraction layer
and a global regressor.
Each local layer is composed of
a local feature extraction layer
and a local regressor.

\subsection{Global layer} \label{sec:globalreg}
The architecture of the global layer is depicted in Figure \ref{fig:archoverview}.b.
The global layer predicts the initial shape estimation $\mathbf{s}^{0}$
from the global feature of image ${I}$.
We use linear regression for this layer
and predict the initial shape $\mathbf{s}^{0}$ directly from the global image features $\boldsymbol{\upphi}^{0}$:

\begin{equation} \label{eq:globalreg}
    \mathbf{s}^{0} =
    \GR (I) = \mathbf{W}^{0} \boldsymbol{\upphi}^{0} + \mathbf{b}^0, \quad \boldsymbol{\upphi}^{0} = g(I)
\end{equation}
where $g(\cdot)$ extracts a $d_{0}$-dimensional global features $\boldsymbol{\upphi}^{0}$
from an image,
$\GR (\cdot)$ represents the global regression function,
$\mathbf{W}^{0} \in \Re^{2P \times d_{0}}$ is the linear regression coefficient matrix,
and $\mathbf{b}^0$ is the bias vector.
For clarity, the later presentation will drop this bias term in the regression function.

The linear regressor with the global feature
gives a coarse estimation of the shape,
which, however, is already a good initialization for the following local layers.

\subsection{Local layer} \label{sec:localreg}
Each local layer refines the shape estimated from the previous layer.
The architecture of the $t$-th local layer is depicted in Figure \ref{fig:archoverview}.b.
It extracts the local (shape-indexed) feature $\boldsymbol{\upphi}^{t}$
and use it to predict the shape increment $\Delta\mathbf{s}^{t}$
using a linear regressor.
The increment is added to $\mathbf{s}^{t-1}$ from the previous layer to produce the refined shape estimation $\mathbf{s}^{t}$.
In mathematical form:


\begin{equation}
    \mathbf{s}^{t} = \LR^{t} (I,\mathbf{s}^{t-1}) = \mathbf{s}^{t-1} + \mathbf{W}^{t} \boldsymbol{\upphi}^{t}, \quad \boldsymbol{\upphi}^{t} = h(I, \mathbf{s}^{t-1}),
    \label{eq:localreg}
\end{equation}

where $h:I, \mathbf{s}^{t-1} \rightarrow \boldsymbol{\upphi}^{t} \in \Re^{d_{t}}$ is the local feature extraction function, $\LR (\cdot,\cdot)$ represents the local regression function,
$\mathbf{W}^{t} \in \Re^{2P \times d_{t}}$ is the linear regression matrix for the $t$-th local regressor.
$\boldsymbol{\upphi}^{t}$ is constructed by concatenating local descriptors around each landmark: $\boldsymbol{\upphi}^{t} = [{\boldsymbol{\upphi}_{1}^{t}}^{\Trans}, {\boldsymbol{\upphi}_{2}^{t}}^{\Trans}, \dots, {\boldsymbol{\upphi}_{P}^{t}}^{\Trans} ] ^{\Trans}$,  $\boldsymbol{\upphi}_{p}^{t}$ is the descriptor extracted around the $p$-th landmark.

Local regressors extract features that describe local appearance and is more suitable for finer adjustment of landmark positions. Besides, it uses only a subset of image pixels for feature extraction and is more advantageous on computational efficiency.



\section{Optimization}
The parameters in the network structure
contain the regression coefficient matrices of the $(T+1)$ regressors:
$\theta = \{\mathbf{W}^0, \mathbf{W}^1, \cdots, \mathbf{W}^T\}$.
These parameters are learned by minimizing the objective function,
$E(\theta) = \frac{1}{2} \sum_{i=1}^N \|DR^T(I_i) - \hat{\mathbf{s}}_i\|_2^2$.
Here $\DR^T(I_i)$ represents the output of the deep regression structure.
It is written from a sequence of sub-network:
$\DR^T(I_i) = \LR^{T}(I_{i}, \DR^{T-1}(I_i))$,
$\DR^t(I_i) = \LR^{t}(I_{i}, \DR^{t-1}(I_i))$,
and $\DR^0(I_i) = \GR(I_{i})$.
We first introduce a sequential learning algorithm that is used
in cascaded regression~\cite{xiong2013supervised}
and empirically show the drawbacks of sequential learning.
Then, we introduce the joint learning algorithm based on back-propagation.

\subsection{Sequential learning} \label{sec:seqlearn}
Sequential learning computes
the regression coefficients
one by one
from $\mathbf{W}^0$ to $\mathbf{W}^T$
to approximately minimize the objective function $E(\theta)$.
The regression coefficient for each regressor
is optimized,
by fixing the trained regression coefficients of the regressors preceding it
and minimizing the difference of its predicted shape
from the true shape.

The coefficient matrix $\mathbf{W}^0$ of the global regressor is solved
as
\begin{equation} \label{eq:seqglobal}
    \mathbf{W}^{0} = \argmin_{\mathbf{W}^{0}} \frac{1}{2N} \sum_{i=1}^{N} \| \mathbf{W}^{0} \boldsymbol{\upphi}_i^0 - \hat{\mathbf{s}}_{i} \|_{2}^{2}.
\end{equation}

The coefficient matrix $\mathbf{W}^t$ of the $t$th local regressor is solved
as
\begin{equation} \label{eq:seqlocal}
    \mathbf{W}^{t} = \argmin_{\mathbf{W}^{t}} \frac{1}{2N} \sum_{i=1}^{N} \| \mathbf{s}^{t-1}_{i} + \mathbf{W}^{t} \boldsymbol{\upphi}^t_{i} - \hat{\mathbf{s}}_{i} \|_{2}^{2},
\end{equation}
where $\boldsymbol{\upphi}_i^t = h(I_i, \mathbf{s}^{t-1}_{i})$
and $\mathbf{s}^{t-1}_{i}$ are fixed
given the coefficients of the first $t$ regressors are estimated.

The sequential learning algorithm is clearly sub-optimal
as the coefficient matrix estimation of each regressor
does not exploit the later regressors.
Empirically,
we observe that
the first few regressors make greater paces
to approach the true shape,
i.e., smaller bias of the shape estimation from those regressors,
while the latter regressors make smaller paces.
Importantly,
we find
that the shape estimation from the first regressors
has larger estimation variances.
This results in the variance of the local (shape-indexed) features is also larger.
As a consequence, it is harder for the later regressors
to make a good shape estimation.

\begin{figure}[t]
\centering
\includegraphics[width=0.9\linewidth]{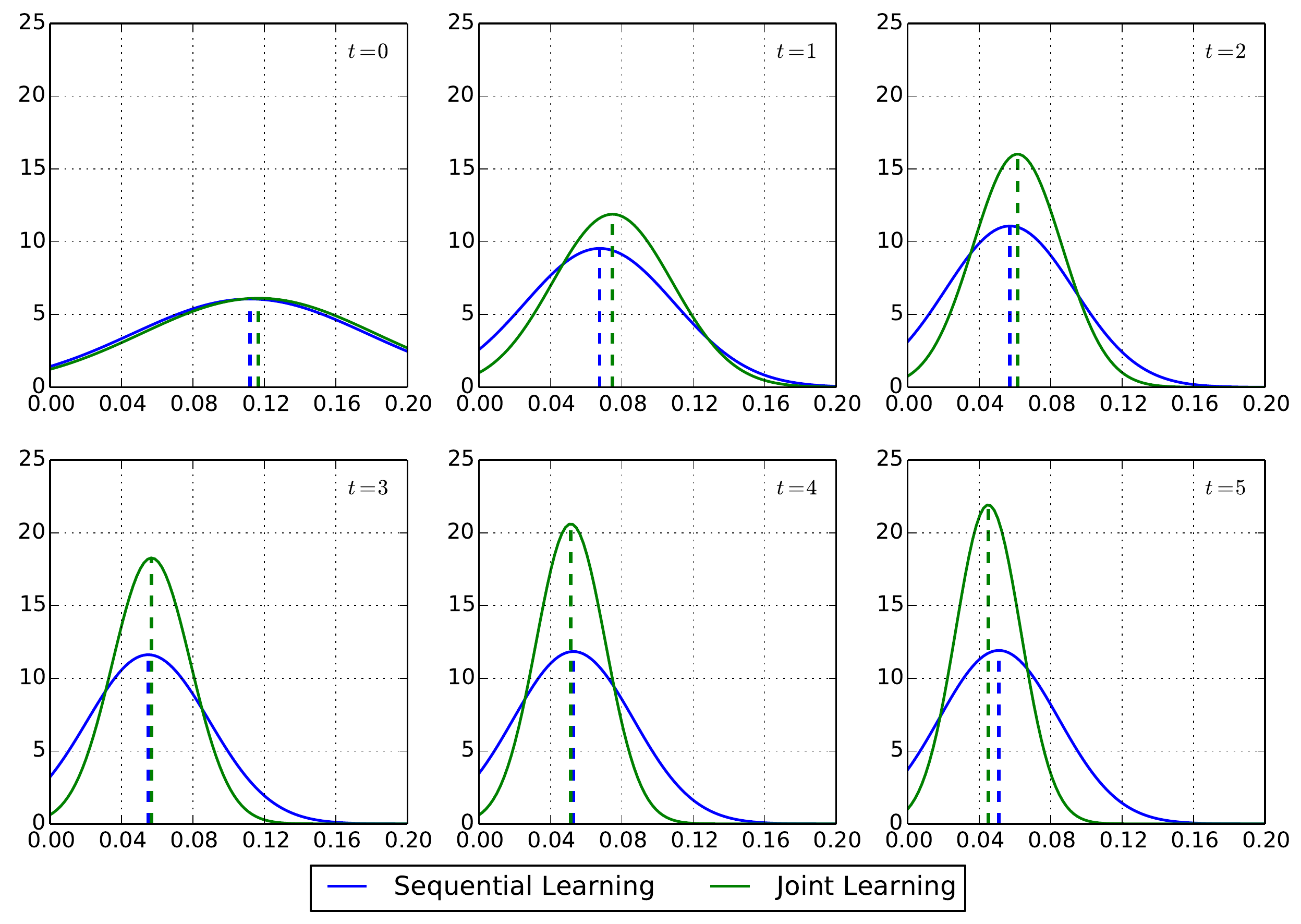}
\caption{Bias and variance comparison of shape estimation error of each stage, learned by sequential and joint learning. Sequential learning over-strongly reduces bias early and results in larger variance (wider and shorter curve), which makes later stages weak. Joint learning balances between bias and variance and makes them gradually and simultaneously decrease, resulting in lower error eventually. The bias and variance are estimated on the 300-W Common Subset and plotted as normal distributions. $x$ axis represents the normalized shape estimation error. (Section \ref{sec:datasets})}
\label{fig:illustrationsOfBiasVariance}
\end{figure}

In the following,
we will introduce the joint learning algorithm using back-propagation
to directly optimize the objective function
such that the optimization of the regression coefficient matrix
helps each other.
The empirical results show that joint learning is able to
make a balanced optimization
of the bias and the variance
of the shape estimation
from the regressors:
both the bias and the variance gradually decrease
from the early regressors to the later regressors.
Consequently, joint learning
yields a better whole shape estimation.
Figure~\ref{fig:illustrationsOfBiasVariance} illustrates
the performance comparison
of each regressors
using sequential learning and joint learning.

\subsection{Joint learning}
We adopt the gradient descent method
to jointly estimate the regression coefficient matrices
by minimizing the global error function $E(\theta)$.
We apply
the back-propagation algorithm \cite{rumelhart1988learning}
to efficiently the evaluate derivatives
of the error function
with respect to the coefficient matrices.

\textbf{The derivatives of local layers.}
The partial derivatives of the error function with respect to $\mathbf{W}^t$ and $\mathbf{s}^t$
are computed using the backward recurrence as:
\begin{align}
\frac{\partial E}{\partial \mathbf{W}^{t}} &= \frac{\partial E}{\partial \mathbf{s}^{t}} \frac{\partial \LR^{t}}{\partial \mathbf{W}^{t}} \\
\frac{\partial E}{\partial \mathbf{s}^{t-1}} & = \frac{\partial E}{\partial \mathbf{s}^{t}} \frac{\partial \LR^{t}}{\partial \mathbf{s}^{t-1}}.
\end{align}


According to Equation \ref{eq:localreg}, $\frac{\partial E}{\partial \mathbf{W}^{t}} =
\boldsymbol{\upphi}^{t}
\frac{\partial E}{\partial \mathbf{s}^{t}}
$.
The partial derivatives $\frac{\partial \LR^{t}}{\partial \mathbf{s}^{t-1}}$ are
computed as:
\begin{equation}
\frac{\partial \LR^{t}}{\partial \mathbf{s}^{t-1}} =
\mathbf{I} + \mathbf{W}^{t} \frac{\partial h}{\partial \mathbf{s}^{t-1}},
\end{equation}
where $\mathbf{I} \in \Re^{2P\times2P}$ is an identity matrix,
and $\frac{\partial h}{\partial \mathbf{s}^{t-1}}$ is the partial derivative
of the local feature extractor with respect to
the shape vector $\mathbf{s}^{t-1}$.

\textbf{The derivatives of local feature extractors.}
The Jacobian matrix of the function $h(I, \mathbf{s})$
with respect to the shape $\mathbf{s}$
is denoted by $\boldsymbol{\uppsi} = \frac{\partial h}{\partial \mathbf{s}}$.
For simplicity, we drop the superscript $t$.
$h(I, \mathbf{s})$ is the local feature extraction operation and is non-differentiable,
and thus its gradients cannot be calculated analytically.
We numerically approximate $\boldsymbol{\uppsi}$ by computing the second-order approximation:
\begin{equation} \label{eq:psiapprox}
\boldsymbol{\uppsi}_{ij}=\frac{\partial\boldsymbol{\upphi}_{i}}{\partial\mathbf{s}_{j}}\approx\frac{h( I,\mathbf{s}^{j+})_{i}-h(I,\mathbf{s}^{j-})_{i}}{2\epsilon},
\end{equation}
where $\mathbf{s}^{j+}$ and $\mathbf{s}^{j-}$ are equal to $\mathbf{s}$ except the $j$-th dimension, where $\mathbf{s}_{j}^{j+}=\mathbf{s}_{j}+\epsilon$ and $\mathbf{s}_{j}^{j-}=\mathbf{s}_{j}-\epsilon$ respectively. The $\epsilon$ is chosen to be a small value which corresponds to several pixels in the image.

Since that $\boldsymbol{\upphi} = [\boldsymbol{\upphi}_{1}^{\Trans},\ \boldsymbol{\upphi}_{2}^{\Trans},\ \dots,\ \boldsymbol{\upphi}_{P}^{\Trans}]^{\Trans}$ is the concatenation of local descriptors extracted around $P$ landmarks, each dimension of $\boldsymbol{\upphi}$ is determined by the corresponding landmarks positions $x_{p},y_{p}$, which are two dimensions of $\mathbf{s}$. Therefore $\boldsymbol{\uppsi}$ is a block-diagonal matrix $\boldsymbol{\uppsi} = \diag ( \boldsymbol{\uppsi}_{1}, \boldsymbol{\uppsi}_{2}, \dots, \boldsymbol{\uppsi}_{P} )$ of which each block $\boldsymbol{\uppsi}_{p} = [ \boldsymbol{\uppsi}_{px}\ \boldsymbol{\uppsi}_{py} ]$. $\boldsymbol{\uppsi}_{px}$ and $\boldsymbol{\uppsi}_{py}$ are respectively the local descriptor gradients along the $x$ and $y$ axis, given by:

\begin{equation}
    \boldsymbol{\uppsi}_{px} = \frac{h_{p}({I},x_{p}+\epsilon, y_{p}) - h_{p}({I},x_{p}-\epsilon, y_{p})}{2\epsilon}
\end{equation}

\begin{equation}
    \boldsymbol{\uppsi}_{py} = \frac{h_{p}({I},x_{p}, y_{p}+\epsilon) - h_{p}({I},x_{p}, y_{p}-\epsilon)}{2\epsilon}
\end{equation}

Here $h_{p}({I},x_{p}, y_{p})$ is the local descriptor function on the $p$-th landmark whose coordinates are $x_{p},y_{p}$.



\textbf{The derivatives of the global layer.}
The derivatives of the error function with respect to
the regression coefficient matrix $\mathbf{W}^0$,
according to Equation \ref{eq:globalreg},
are computed as follows:
\begin{equation}
    \frac{\partial E}{\partial \mathbf{W}^{0}} =
    \frac{\partial E}{\partial \mathbf{s}^{0}} \frac{\partial \GR}{\partial \mathbf{W}^0}
    = \boldsymbol{\upphi}^{0} \frac{\partial E}{\partial \mathbf{s}^{0}}.
\end{equation}

\subsubsection{Pre-training and dropout}

In order to obtain a good initialization for joint optimization,
we pre-train the network by sequential learning.
As we use the dropout strategy for joint optimization,
we use the gradient decent algorithm with the dropout strategy
to estimate the regression coefficient matrices
to solve Equations \ref{eq:seqglobal} and \ref{eq:seqlocal}.

The dropout technique \cite{hinton2012improving} has been shown helpful in deep neural network training,
being a strong and adaptive regularizer.
We adopt this technique to joint learning,
which is critical
to avoid over-fitting.
During the forward propagation, each dimension of features $\boldsymbol{\upphi}^{t}$ is set to zero with probability $1-p$.
In back propagation the gradients on those dimensions are also set to zero. For local layers, the forward and backward propagation process are given by:
$
    \mathbf{s}^{t} = \mathbf{s}^{t-1} + \mathbf{W}^{t} \mathbf{D}_{\mathbf{z}}^{t} \boldsymbol{\upphi}^{t}
$
and
$
    \frac{\partial E}{\partial \mathbf{s}^{t-1}} = \frac{\partial E}{\partial \mathbf{s}^{t}} (\mathbf{I} + \mathbf{W}^{t} \mathbf{D}_{\mathbf{z}}^{t} \boldsymbol{\uppsi}^{t}),
$
respectively. Here $\mathbf{D}_{\mathbf{z}}^{t} = \diag (\mathbf{z}) \in \Re^{d_{t} \times d_{t}}$, diagonal elements $z_{i}$ are sampled from a Bernoulli distribution $z_{i} \sim \textrm{Bernoulli} ( p )$. During test, $\mathbf{W}^{t} \mathbf{D}_{\mathbf{z}}^{t}$ is replaced by $p \mathbf{W}^{t}$. The probability $p$, or dropout rate, is fixed to $0.5$ throughout our experiments. For the global layer, the dropout is done in a similar way.

\subsubsection{Implementation details}

For global features $g({ I})$ we use the HOG \cite{dalal2005histograms} descriptor. Descriptors are computed on images down-sampled to sizes of $64 \times 64$.
Block size, block stride, cell size and number of bins are chosen as $16 \times 16$, $16 \times 16$, $8 \times 8$ and $9$ respectively.
This results in global features with $1764$ dimensions.
For local features $h({ I},\mathbf{s})$, we use a modified version of the SIFT descriptor \cite{lowe2004distinctive}.
$128$-d descriptors are extracted around each landmark, and concatenated to produce local features $\boldsymbol{\upphi} \in \Re^{128P}$.
Since the numerical approximation of $\boldsymbol{\uppsi}$ requires a great number of feature extraction operations and storing SIFT descriptors on all image locations requires too much memory, we modify the original SIFT descriptor so that it can be computed faster. For each image, the responses for $8$ orientation bins on all locations are pre-computed and stored in $8$ response maps. The Gaussian weight mask is dropped and the spatial bin interpolation is implicitly approximated by blurring the response maps using a Gaussian kernel. This is inspired by the DAISY descriptor \cite{tola2010daisy}. After that the response maps are converted to integral maps, where histograms can be computed with only a few addition and substraction operations \cite{porikli2005integral}. The response maps are pre-computed and stored in memory, so that the descriptors can be efficiently extracted during running time.

For both datasets, we set the number of local layers to $T = 5$.
SIFT patch sizes for the first $4$ local layers are set to $32 \times 32$.
The last two local layers have smaller sizes of $16 \times 16$.
We augment training samples by flipping training images horizontally.
A validation set of $200$ samples is split out from the training set for monitoring the training process.

The $\epsilon$ in Equation \ref{eq:psiapprox} is set to $2$ pixels throughout our experiments. Other small values have also been tried but have no significant impact.
Network parameters are updated by Stochastic Gradient Descent \cite{lecun2012efficient} with momentum \cite{sutskever2013importance} set to $0.9$.
The mini-batch size is set to $100$.
During training, the learning rate is set to $10^{-2}$ at first and manually decreased when validation error stops to decrease \cite{krizhevsky2012imagenet}.
The training process is terminated when the validation error stops to decrease for enough number of iterations.


\section{Experiments} \label{sec:exp}

\subsection{Datasets and evaluation metric} \label{sec:datasets}

\textbf{Datasets:} Performance is evaluated on the LFPW dataset \cite{belhumeur2011localizing} and the 300-W dataset \cite{sagonas2013semi}. The LFPW dataset is annotated by $29$ landmarks. The dataset provides URLs only and some are no longer valid. We use $717$ of the $1100$ images for training and $249$ of the $300$ images for testing.
300-W dataset is created from several re-annotated datasets including LFPW \cite{belhumeur2011localizing}, AFW \cite{zhu2012face}, Helen \cite{le2012interactive} and XM2VTS \cite{messer1999xm2vtsdb}. The number of landmarks is $68$. Since the official testing set of 300-W is not publicly available, we follow \cite{renface} and build the training set using AFW, the training set of LFPW and the training set of Helen, with $3148$ images in total.
Our testing set consists of IBUG, the testing set of LFPW and the testing set of Helen, with 689 images in total.
Also following \cite{renface}, we evaluate performance on 1) all images of the testing set, called the Fullset 2) testing sets of Helen and LFPW, called the Common Subset and 3) IBUG dataset, called the Challenging Subset.

\def\arraystretch{1.2}
\def\MyColumnWidth{1.5cm}
\begin{table}
    \caption{Results on the LFPW and the 300-W datasets, measured by the shape error normalized by the inter-pupil distance. *The original SDM and ESR paper does not include results on the 300-W dataset and we quote results from \cite{renface}.}
    \begin{center}
    \begin{tabular}{l >{\centering\arraybackslash}m{\MyColumnWidth} >{\centering\arraybackslash}m{\MyColumnWidth} >{\centering\arraybackslash}m{\MyColumnWidth}}
    \multicolumn{2}{c}{\textbf{LFPW}} \\
    \hline
    Method & Normalized Error \\
    \hline
    CoE \cite{belhumeur2011localizing} & 3.90 \\
    ESR \cite{cao2014face} & 3.47 \\
    RCPR \cite{burgos2013robust} & 3.50 \\
    SDM \cite{xiong2013supervised} & 3.47 \\
    LBF \cite{renface} & \textbf{3.35} \\
    \hline
    SequentialReg & 3.90 \\
    DeepRegLocal & 3.47 \\
    DeepReg & 3.45 \\
    \hline
    \end{tabular}
    \quad \quad
    \begin{tabular}{l >{\centering\arraybackslash}m{\MyColumnWidth} >{\centering\arraybackslash}m{\MyColumnWidth} >{\centering\arraybackslash}m{\MyColumnWidth}}
    \multicolumn{4}{c}{\textbf{300-W}} \\
    \hline
    Method & Fullset & Common Subset & Challenging Subset \\
    \hline
    \\
    ESR* \cite{cao2014face} & 7.58 & 5.28 & 17.00 \\
    \\
    SDM* \cite{xiong2013supervised} & 7.52 & 5.60 & 15.40 \\
    LBF \cite{renface} & 6.32 & 4.95 & \textbf{11.98} \\
    \hline
    SequentialReg & 7.31 & 5.11 & 16.35 \\
    DeepRegLocal & 6.57 & 4.67 & 14.30 \\
    DeepReg & \textbf{6.31} & \textbf{4.51} & 13.80 \\
    \hline
    \end{tabular}
    \end{center}
    \label{tbl:resultsComparison}
\end{table}

\textbf{Evaluation metric:} Following \cite{belhumeur2011localizing}, we evaluate performance by the average landmark error normalized by inter-pupil distance:

\begin{equation} \label{eq:evalmetric}
    \textrm{error} = \frac{1}{N}\sum_{i=1}^{N} \frac{\frac{1}{P}\sum_{p=1}^{P}\sqrt{(x_{p}^{(i)}-\hat{x}_{p}^{(i)})^{2}+(y_{p}^{(i)}-\hat{y}_{p}^{(i)})^{2}}}{d_{\textrm{pupils}}^{(i)}},
\end{equation}

where $\hat{x}_{p}^{(i)},\hat{y}_{p}^{(i)}$ are ground truth coordinates for the $p$-th landmark of the $i$-th sample, $d_{\textrm{pupils}}^{(i)}$ is the inter-pupil distance of the $i$-th sample. For 300-W, pupil landmarks are not annotated and are replaced by the mean landmarks of the landmarks around each eye.

\subsection{Evaluation}

\begin{figure}[t]
    \begin{centering}
    \includegraphics[width=\linewidth]{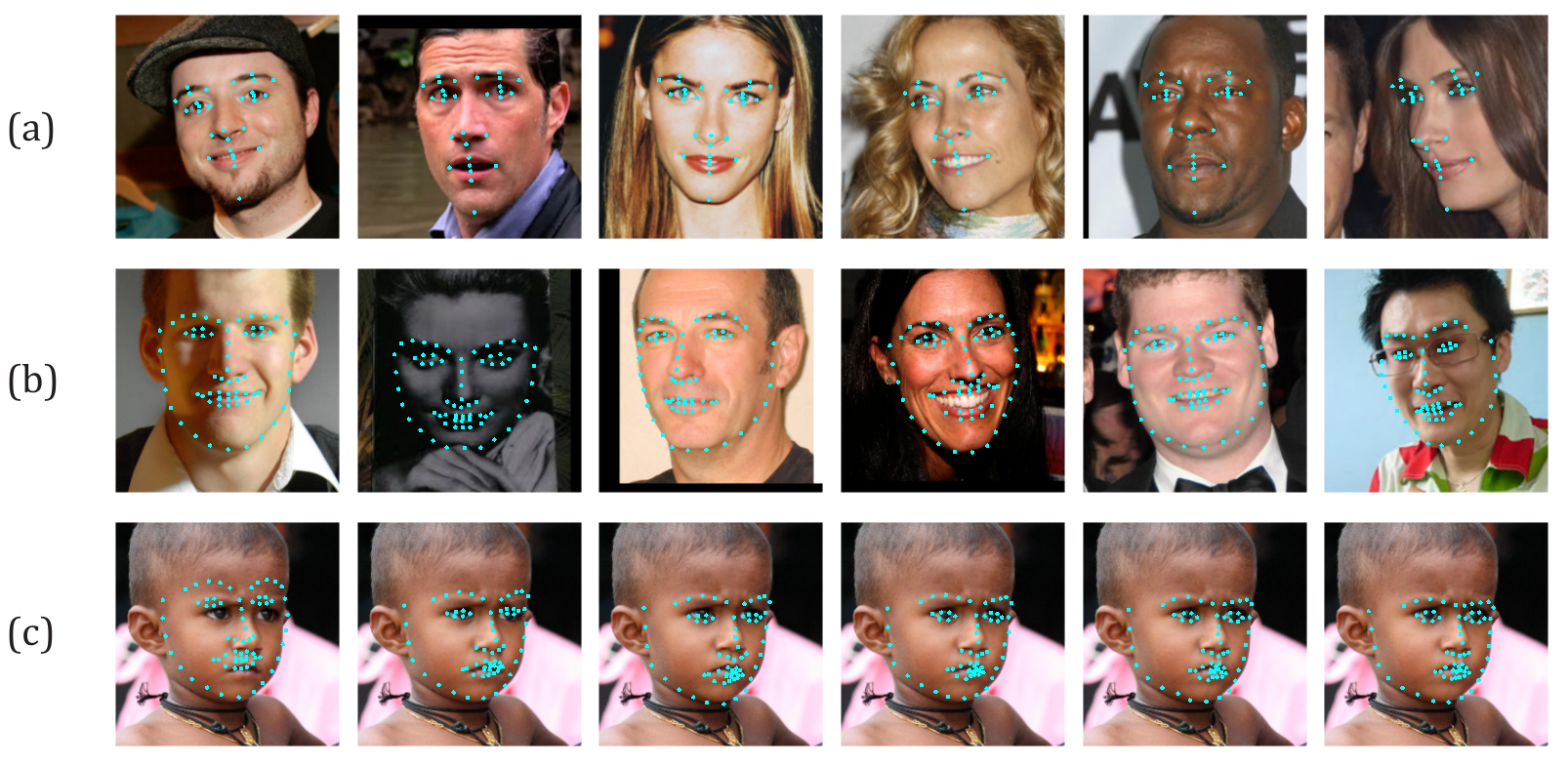}
    \par\end{centering}
    \caption{(a) Example results on the LFPW dataset. (b) Example results on the 300-W dataset. (c) Estimation by each stage from $t=0$ to $5$.}
    \label{fig:resultvis}
\end{figure}

We term our approach as~\emph{DeepReg},
our approach with sequential learning as~\emph{SequentialReg}
and a variant of the network which drops the global regressor as~\emph{DeepRegLocal}.
The initial shape estimation $\mathbf{s}^{0}$ in~\emph{DeepRegLocal} is given
by the mean shape $\bar{\mathbf{s}}$ calculated from the training set,
as adopted in cascaded regression~\cite{xiong2013supervised,burgos2013robust,renface}.
First, we compare the result of~\emph{DeepReg} with
the two baseline algorithms:~\emph{SequentialReg} and~\emph{DeepRegLocal}.
The results are listed in Table~\ref{tbl:resultsComparison} and visualized in Figure~\ref{fig:resultvis}.

One can see from Table~\ref{tbl:resultsComparison} that
\emph{DeepReg} outperforms both~\emph{SequentialReg} and~\emph{DeepRegLocal}.
The superiority over~\emph{SequentialReg}
stems from joint optimization,
which is able to balance the biases and the variances
of all the regressors.
The superiority over~\emph{DeepRegLocal}
is because the global regressor is helpful to generate
a robust initialization.
Second,
in comparison with
the closely-related regression algorithm,
supervised descent method (SDM, \cite{xiong2013supervised}),
our approach performs better.
The superiority of our approach and~\emph{DeepRegLocal}
is not as significant as that to~\emph{SequentialReg}.
The reason is that SDM did good job on feature transform,
which potentially can help our approach,
for example, including feature transform and even convolutions into our deep regress framework.
Last,
we also report the comparison results
with
other state-of-the-art algorithms,
including the algorithm using a consensus of exemplars~(CoE~\cite{belhumeur2011localizing}),
explicit shape regression (ESR \cite{cao2014face}),
robust cascaded pose regression (RCPR \cite{burgos2013robust})
and local binary features (LBF \cite{renface}),
in which our approach and LBF perform the best.
As shown in Table \ref{tbl:resultsComparison},
our approach performs better in 300-W Fullset and 300-W Common Subset
over LBF,
but poorer in LFPW and 300-W Challenging Subset.
The reason is that LBF performs an extra feature learning step
that is essential for good performance,
and in essence our approach is able to benefit from this step
if we can reproduce their reported results.

\subsection{Empirical analysis} \label{sec:seqvsjt}

\begin{figure}[t]
    \begin{centering}
    \includegraphics[width=\linewidth]{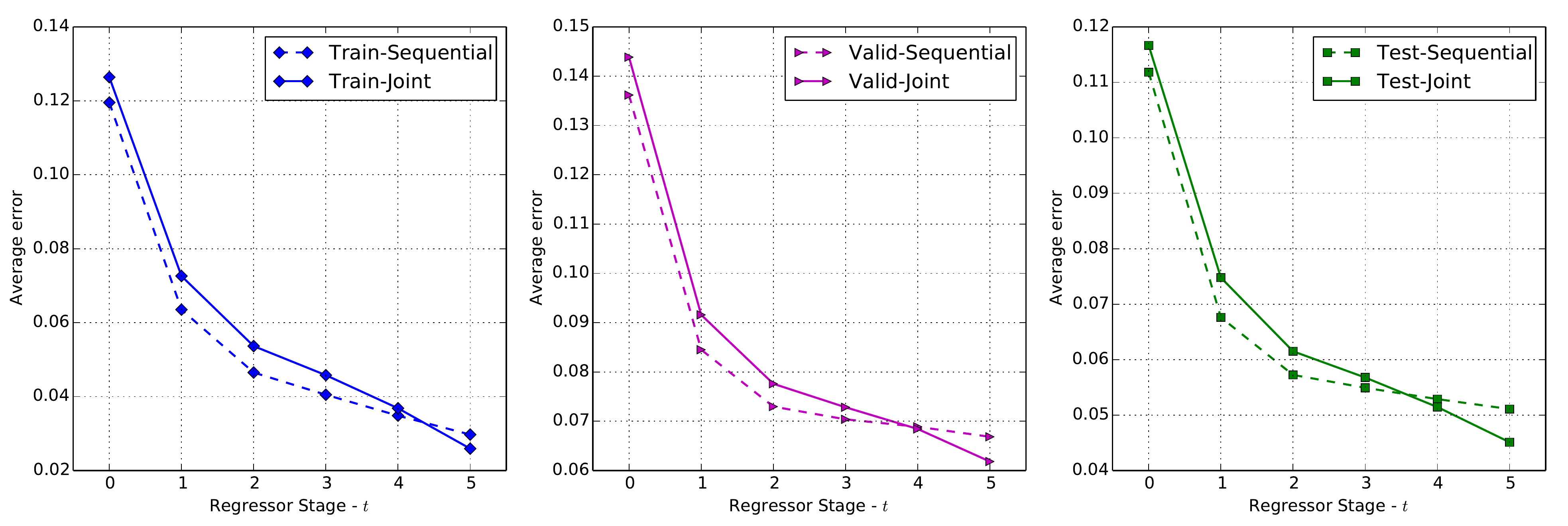}
    \par\end{centering}
    \caption{Average estimation error of all stages, on training, validation and testing set. Dashed line represents the sequentially trained layers and the solid line represents jointly trained layers (Tested on the 300-W dataset).}
    \label{fig:stageerrs}
\end{figure}

Figure \ref{fig:stageerrs} plots the estimation errors of all stages on the training, validation and testing sets. One can see from the plot that sequential learning tends to result in strong early stages which eliminate most of the error. The later stages, however, are much weaker.
Joint learning mitigates this and
the estimation gradually and evenly approaches the ground truth,
resulting in a flattened error curve and better estimation eventually.
Furthermore, as shown in Figure \ref{fig:illustrationsOfBiasVariance}, joint learning balances between bias and variance and makes them
gradually and simultaneously decrease,
while in sequential learning
the variance decreases much slower.




\section{Conclusion}

In this paper,
we present a deep regression approach to face alignment.
We adopt back-propagation with the dropout strategy
to jointly optimize the regression coefficient matrices
of a deep network,
a sequence of one global linear regressor
and multi-stage local regressors.
The benefit of joint optimization
lies in that the resulting regressors gradually and simultaneously decrease
the bias and the variance of each shape estimator
and make harmonious contributions to shape prediction,
yielding a superior shape predictor over
the sequential learning algorithm as done in cascaded regression.
Experimental results demonstrate the powerfulness of the proposed approach.

\subsubsection*{Acknowledgements}
This work was partially supported by National Natural
Science Foundation of China (NSFC) (No. 61222308), and
in part by NSFC (No. 61173120), Program for New Century Excellent Talents in University (No. NCET-12-0217)
and Fundamental Research Funds for the Central Universities (No. HUST 2013TS115).

\subsubsection*{References}

{
\small{}
\bibliographystyle{plain}
\renewcommand{\section}[2]{}
\bibliography{references}
}

\end{document}